\documentclass[11pt]{article}

\usepackage[preprint]{acl}
\usepackage{format/custom_package}
\usepackage{format/macro}

\title{\sysname: Can LLMs Tell When Simulation Errors\\
Matter in AI-Driven Systems Research?}

\author{Jinwoo Hwang\textsuperscript{1*} \quad Sukmin Cho\textsuperscript{1*} \quad Tao Zhang\textsuperscript{2} \\
\textbf{Soheil Abbasloo\textsuperscript{2} \quad Jongse Park\textsuperscript{1\dag} \quad Peng Cheng\textsuperscript{2}} \\
\textsuperscript{1}KAIST \quad \textsuperscript{2}Microsoft Research Asia}

\begin{document}

\maketitle
{\makeatletter\renewcommand{\@makefntext}[1]{\noindent#1}\makeatother\renewcommand{\thefootnote}{}%
\footnotetext{\raggedright\textsuperscript{*}Work done during an internship at Microsoft Research Asia.\\\textsuperscript{\dag}Corresponding author.}}

\begin{abstract}
LLM agents increasingly drive systems research and rely on simulators to evaluate candidate designs at scale.
Simulators are imperfect by design: their modeling limitations can distort predictions yet leave the recommended design correct, or change which design should be selected.
We introduce \sysname to test whether LLMs can tell these cases apart, judging from their systems understanding when a limitation invalidates the simulator's recommendation.
By perturbing a simulator, \sysname constructs workload pairs with the same candidates, limitation, and simulator recommendation.
The perturbation changes predictions on both workloads, yet a reference simulator supports the recommendation on only one.
Evaluating thirteen frontier LLMs on LLM serving design problems, we find that identifying a limitation does not ensure a correct choice, and even the strongest models sometimes abandon valid recommendations.
Self-correction through a review turn helps these models restore valid choices but introduces new errors in several others.
These findings motivate evaluating systems understanding through design decisions under imperfect simulators.
\end{abstract}

\section{Introduction}
\label{sec:introduction}

LLM agents are opening a new paradigm for research, in which agents themselves propose ideas, run experiments, and iterate on tasks that would take a skilled researcher days~\cite{openai-research-intern:misc:2026,automated-w2s:misc:2026}.
In systems research, AI-Driven Research for Systems (ADRS) realizes this paradigm through an iterative process in which LLMs generate candidate designs, evaluate their performance, and refine them using the resulting feedback~\cite{barbarians-in:arxiv:2025}.
The same process now drives data-center scheduling, serving policies, load balancing, and database optimization~\cite{alphaevolve:arxiv:2025,glia:cais:2026,shinkaevolve:iclr:2026,adrs-databases:arxiv:2026}.
As LLMs take on more of these steps, they must also interpret evaluation results and decide which candidate designs to retain or revise.

\begin{figure}[t]
\centering
\includegraphics[width=\columnwidth]{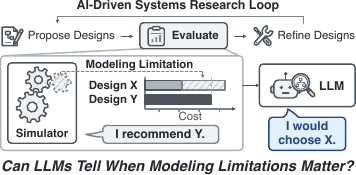}
\caption{Simulator-assisted evaluation in ADRS. A modeling limitation can invalidate its recommendation.}
\label{fig:concept}
\end{figure}

Simulators support these evaluations with a software model of a computer system that predicts how a candidate design performs under a given workload.
This makes evaluation tractable without deploying candidates on real hardware~\cite{vidur:mlsys:2024,barbarians-in:arxiv:2025,adrs-databases:arxiv:2026}.
To remain scalable, simulators model system behavior at a chosen level of detail~\cite{simulation-detail:scs:2001}.

The details that simulators approximate or omit introduce \textit{modeling limitations}.
For example, ignoring competition for shared network bandwidth lets a simulator scale to thousands of accelerators but can underestimate transfer delays~\cite{astra-sim2:ispass:2023}.
Such a limitation may alter predicted performance while leaving the best candidate unchanged, or it may change which candidate to select for a particular workload~\cite{song:wsc:2015}.
Using a simulator well therefore requires judging the \textit{decision impact} of its limitations, that is, whether they matter for the decision at hand~\cite{network-simulation-validation:ieee-network:2001}.
As LLMs take over these decisions, they must make this judgment themselves, yet how well they do so remains understudied.

\Cref{fig:concept} illustrates such a case, in which a cost that the limitation adds reverses the ranking of two designs and thereby invalidates the simulator's recommendation.
We argue that this judgment is the form of systems understanding that ADRS relies on: an LLM need not predict a system's performance on its own, but it must recognize when a simulator of that system can be trusted for a decision.
Existing evaluations do not isolate this judgment: whether a simulator's modeling limitations invalidate its recommendation for a given design decision.
We therefore ask three questions:
\begin{itemize}[nosep]
    \item \textbf{RQ1}: How does simulator evidence shape the design decisions of LLMs?
    \item \textbf{RQ2}: Can LLMs judge the decision impact of a simulator's modeling limitations?
    \item \textbf{RQ3}: Can LLMs improve this judgment via their self-correction capability?
\end{itemize}

To answer these questions, we introduce \sysname, an LLM evaluation framework for simulator-assisted system design review.
\sysname mimics the modeling limitations of real-world simulators by injecting controlled perturbations, small code changes to how the simulator represents work, costs, or resource interactions.
For each perturbation, \sysname pairs two workloads on which the perturbed simulator recommends the same design.
The reference simulator supports that recommendation on one workload (\textit{upheld}) but favors another design on the other (\textit{overturned}).

\sysname operationalizes \textit{judgment} as correct design selection across upheld and overturned problems.
It measures judgment by overall accuracy, the mean of the accuracies on the two problem types.
A fixed candidate preference, or a rule that always follows or always rejects the presented recommendation, solves at most one problem of each pair.
\textit{Disposition} summarizes the imbalance between the two accuracies.
We measure judgment and disposition at every level of simulator evidence and compare them before and after a review turn that elicits self-correction.
Our contributions are as follows:
\begin{itemize}[nosep]
    \item We argue for evaluating systems understanding in ADRS through judgments of when simulator limitations invalidate a design recommendation.
    \item We propose \sysname, which introduces modeling limitations through controlled perturbations and measures the judgment and disposition of LLMs on paired upheld and overturned problems.
    \item Evaluating thirteen frontier-level LLMs, we find that most LLMs show limited judgment even when they identify the limitation, and that inspection instructions and self-correction often shift disposition more than they raise judgment.
\end{itemize}

\section{Background}
\label{sec:background}

\begin{table*}[t]
\centering
\footnotesize
\begin{tabular*}{\textwidth}{@{\extracolsep{\fill}}llll@{}}
\toprule
Simulator & Omitted or approximated behavior & Simulator & Omitted or approximated behavior \\
\midrule
APEX & Tokenization and position embeddings & FlexFlow & Runtime overhead, full-bandwidth transfers \\
Vidur & Per-request context lengths as total KV reads & TopoOpt & Packet behavior during search \\
LLMServingSim & Per-block compilation, GPU kernel tuning & SimGrid & TCP slow start, packet-level behavior \\
ASTRA-sim 2.0 & Congestion, in the analytical network model & FireSim & Storage timing, as a functional block device \\
GenZ & Microarchitecture and expert-load imbalance & BookSim2 & Full-system traffic, in trace-replay mode \\
\bottomrule
\end{tabular*}
\caption{Examples of documented modeling choices from APEX~\cite{apex:arxiv:2025}, Vidur~\cite{vidur:mlsys:2024}, LLMServingSim~\cite{llmservingsim:iiswc:2024}, ASTRA-sim 2.0~\cite{astra-sim2:ispass:2023}, GenZ~\cite{genz:arxiv:2025}, FlexFlow~\cite{flexflow:mlsys:2019}, TopoOpt~\cite{topoopt:nsdi:2023}, SimGrid~\cite{simgrid-network:valuetools:2007}, FireSim~\cite{firesim:isca:2018}, and BookSim2~\cite{booksim2:ispass:2013}.}
\label{tab:abstractions}
\end{table*}

\subsection{Simulator-Assisted Systems Research}
\niparagraph{Simulation-based evaluation.}
In systems research, a simulator is an executable model used to estimate how a design performs under a specified workload.
The workload describes the work to process, while the model represents relevant system components and their interactions.
For example, researchers can compare routing algorithms under the same network traffic by simulating packet movement and resource contention and comparing the predicted latency and throughput~\cite{booksim2:ispass:2013}.
This allows candidate designs to be evaluated without deploying each on the target system.

\niparagraph{Simulator use in ADRS.}
ADRS systems such as AlphaEvolve~\cite{alphaevolve:arxiv:2025} and ShinkaEvolve~\cite{shinkaevolve:iclr:2026} run LLM-driven program search, in which evaluator scores select the candidates that seed subsequent changes.
In simulator-assisted ADRS, this evaluator is a simulator, so its results determine how each design is assessed.
The same results also guide which designs LLM agents explore next.
Glia~\cite{glia:cais:2026} refines designs by analyzing simulation experiments, and Database ADRS~\cite{adrs-databases:arxiv:2026} goes further by co-evolving a simulator with candidate policies in its buffer-cache study.

\subsection{Modeling Limitations and Decisions}
\niparagraph{Modeling limitations.}
Simulators keep evaluation tractable by modeling only the detail that their designers judge relevant to the research question~\cite{simulation-detail:scs:2001}.
\Cref{tab:abstractions} lists examples of such choices that simulator authors document, including omitted operations, summarized request state, simplified surrounding components, and idealized load or communication.

\niparagraph{Decision impact.}
Such choices matter to a design decision only through their effect on the comparison between candidates.
Ranking-and-selection analyses show that an equal additive error across alternatives cancels in pairwise differences, whereas candidate-dependent errors can widen, narrow, or reverse a performance gap~\cite{song:wsc:2015}, and architecture studies warn that relative trends may not survive simulator errors that interact with the evaluated technique~\cite{simulators-harmful:ieee-micro:2015}.
Documented examples include errors that distort performance gaps~\cite{simulation-detail:scs:2001} and errors that reverse design rankings~\cite{apex:arxiv:2025}.
Judging decision impact therefore requires connecting a limitation's effect on costs and resource interactions to the specific candidates and workload.

\subsection{Agents under Unreliable Evidence}
\niparagraph{Imperfect external evidence.}
Prior work examines how unreliable evidence affects LLMs.
GARAG~\cite{garag:findings-emnlp:2024} searches for low-level document perturbations that affect retrieval and answer generation in RAG pipelines.
\citet{dont-blindly-trust:arxiv:2026} show that persistent misleading feedback can underperform a matched no-feedback condition, while \citet{tool-overreliance:arxiv:2026} show that agents can recognize a conflict yet still adopt corrupted tool content.
\sysname shares this concern with the gap between critique and action, examining it via controlled perturbations of simulator computations.

\niparagraph{Simulation-grounded decisions.}
MechSim~\cite{mechsim:arxiv:2026} has agents reason over the mechanisms and assumptions of simplified candidate simulators and test changes through re-execution, scoring their policy rankings in aggregate against a more detailed oracle simulator.
\citet{controller-judgment:processes:2026} study how LLMs rank fixed candidate controllers from simulation results under different information conditions.
\sysname isolates whether a particular modeling limitation invalidates a recommendation, scoring upheld and overturned problems separately.
It pairs workloads that share the candidate designs, the limitation, the metric, and the simulator's recommendation, while the reference supports that recommendation on only one workload.

\niparagraph{Appropriate reliance.}
Appropriate reliance involves following correct advice and rejecting incorrect advice~\cite{appropriate-reliance:iui:2023}.
Tools Fail~\cite{tools-fail:emnlp:2024} treats outputs with task-irrelevant errors as acceptable and finds that models often reject them.
\sysname examines a related distinction between computational accuracy and recommendation validity, since an upheld problem contains a limitation that changes predictions while the reference prefers the recommended design.

\section{\sysname Design}
\label{sec:methodology}
\label{sec:problem}

\begin{figure*}[t]
\centering
\includegraphics[width=\textwidth]{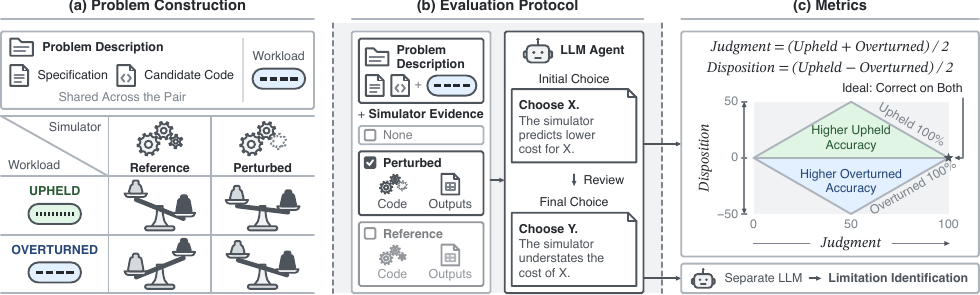}
\caption{Overview of \sysname, from constructing paired problems with a perturbed simulator to scoring LLM choices.}
\label{fig:overview}
\end{figure*}

\Cref{fig:overview} summarizes \sysname, which tests whether an LLM can tell when a modeling limitation invalidates a simulator's recommendation.
An LLM selects a design for a workload from fixed candidates.
We construct upheld and overturned problem pairs, vary the simulator evidence the LLM reads, and collect an initial choice followed by a review.
Comparing both choices with the reference simulator's answer yields judgment and disposition, while a separate annotation records whether each response identifies the modeling limitation.

\subsection{Problem Construction}
\label{sec:task-construction}

\niparagraph{Problem format.}
Each \sysname problem is a multiple-choice question whose options are candidate designs.
The question asks which candidate achieves the best target metric for a given workload, and its reference answer is the design the reference simulator predicts best under the fixed system specification.
The context is a set of files the LLM reads with read-only tools: the \textit{problem description}, namely the system specification, candidate code, and workload, plus simulator evidence set by the evidence condition.
Each problem is built on a perturbed simulator, a copy of the reference simulator with one changed computation.
The LLM selects among the given candidates without writing a new design or repairing the simulator.

\niparagraph{Perturbations.}
Each perturbation is an explicit program change to a simulator computation that every candidate relies on, adapted from classic mutation operators~\cite{mutation-testing:tse:2011,mutation-advances:adcom:2019}, that inserts a modeling limitation, any discrepancy in how the simulator represents work, costs, or resource interactions.
\sysname admits a change only if varying the workload can make the recommendation correct on some workloads and wrong on others, and if it is reviewable against the specification and the files the LLM reads.
\Cref{tab:families} in the appendix lists the perturbations.

\niparagraph{Upheld and overturned pairs.}
LLM agents separate from the evaluated LLMs build the problems: a builder proposes workloads, and reviewers check the evidence, as \cref{sec:protocol-details} describes.
For each proposed workload, \sysname runs every candidate under both simulator versions and records their recommendations and whether the change alters predictions.
A problem is \textit{upheld} when both versions recommend the same design and \textit{overturned} when they differ, and the perturbed simulator's recommendation is the presented recommendation.

A \textit{perturbation pair} joins an upheld and an overturned problem that share the candidates, perturbation, target metric, and presented recommendation and differ only in workload, and in both the changed computation alters predictions.
\Cref{fig:overview}(a) depicts one such pair.
Such pairs arise because candidates differ in how strongly the change affects them, while the workload sets how much that difference matters.
This contrast tests whether an LLM distinguishes a limitation's presence from its impact on the design choice, since neither identifying the limitation nor a fixed preference, such as always following, can solve both problems.

\niparagraph{Validation.}
Independent solvers given only the problem description and the perturbed simulator's evidence attempt each problem, and an evidence audit checks whether these files support the reference answer.
\sysname drops each failing problem with its pair partner and fixes every problem and pair before evaluation.
This check does not ensure that the problem description alone suffices.

\subsection{Evaluation Protocol}
\label{sec:evaluation-protocol}
We compare evidence conditions on the same problems to study evidence use, and initial with reviewed choices to study self-correction.

\niparagraph{Evidence conditions.}
In \Cref{fig:overview}(b), an evidence condition adds simulator evidence beyond the problem description, from none to all of the simulator source, the measured or specified cost parameters it evaluates, the modeled system's configuration, its predictions, and its execution records.
The no-simulator condition adds none, the perturbed condition adds all of the perturbed simulator's evidence, and the reference condition adds all of the reference simulator's evidence as a control.
All conditions of a problem are scored against the same reference answer.
Each conversation covers one problem under one condition, so the LLM never sees the pair partner or label, and in the perturbed condition it sees no reference results and is not told what changed.

\niparagraph{Two-turn review.}
In both turns, the LLM can read the problem's files with read-only tools.
The first request asks only for a recommendation with a short rationale and does not mention modeling limitations or the review that follows, so the initial choice reflects the LLM's own reading of the evidence.
The second request then asks the LLM to review that choice and name the modeling limitation, if any, that it considers consequential.
No new evidence or correctness feedback arrives between the two turns, although the LLM may open files it skipped in the first turn.
\Cref{fig:responses} shows a conversation, and \cref{sec:mechanism-diagnostic-protocol} gives the full protocol.

\subsection{Metrics}
\label{sec:scoring}
\niparagraph{Judgment and disposition.}
For each model, evidence condition, and turn $t$, let $U_t$ and $O_t$ denote accuracy on upheld and overturned problems.
A choice is correct only if it matches the reference answer, so selecting a third design on an overturned problem remains incorrect.
\sysname summarizes judgment by the mean of the two accuracies and disposition by their imbalance:
\begin{equation}
J_t = \frac{U_t + O_t}{2}, \qquad D_t = \frac{U_t - O_t}{2}.
\label{eq:judgment-disposition}
\end{equation}
In percent, $J_t$ ranges from 0 to 100 and $D_t$ from $-50$ to $50$.
Positive $D_t$ indicates higher upheld accuracy and negative $D_t$ the reverse, as in \Cref{fig:overview}(c).
This imbalance differs from the rate of following the recommendation, since an override may select an incorrect alternative.
Errors that keep an invalid recommendation, abandon a valid one, or select a third design are counted separately.

\niparagraph{Review and identification.}
Review changes are summarized by $\Delta J = J_{\mathrm{final}} - J_{\mathrm{initial}}$ and $\Delta D = D_{\mathrm{final}} - D_{\mathrm{initial}}$, together with counts of corrected and newly incorrect choices.
Every assigned problem is scored, and a missing or invalid choice is incorrect.
Separately, \textit{limitation identification} records whether a response identifies the modeling limitation that the problem's perturbation introduces, such as transfers modeled without sharing a GPU's network interface.
A separate LLM annotates each perturbed response with a fixed rubric, blind to labels, reference answers, and model identity, and identification earns no credit for the design choice, as \cref{sec:scoring-analysis} details.

\section{Experimental Setup}
\label{sec:case-study}

\begin{figure}[t]
\convturn{First turn}
\begin{convuser}
\ldots\ Recommend one candidate for this workload.
\end{convuser}
\begin{convtool}{Tool call 2 of 64}
read prediction.json\par\smallskip
{\normalfont\scriptsize\itshape Predicted p95 end-to-end latency (s), values only:}\\
A 2.424 \quad B 2.425 \quad C 1.816 \quad D 1.649
\end{convtool}
\begin{convtool}{Tool call 29 of 64}
read \ldots/systems/multi\_instance.rs:179-250
\convadd{+ network\_sim.add\_link(source, destination, \ldots}
\end{convtool}
\begin{convagent}{Initial choice: D}
``\ldots\ Candidate D achieves the lowest 95th-percentile end-to-end latency of \textasciitilde1.649s.''
\end{convagent}
\convturn{Review turn}
\begin{convuser}
Review your recommendation before it is final. \ldots
\end{convuser}
\begin{convagent}{Final choice: C}
``\ldots\ those concurrent transfers would share 1.25 GB/s rather than each getting full bandwidth \ldots\ pushing it to \textasciitilde2.02s and making Candidate C (1.816s, zero network transfers) the faster policy in reality.''
\end{convagent}
\begin{conveval}
\convlabel{Evaluator only}\quad Reference answer: C. Outcome: overturned.
\end{conveval}
\caption{An example \sysname conversation, abridged from \Cref{sec:case-handoff}. The LLM never sees the last box.}
\label{fig:responses}
\end{figure}

\begin{figure*}[t]
\centering
\includegraphics[width=\textwidth]{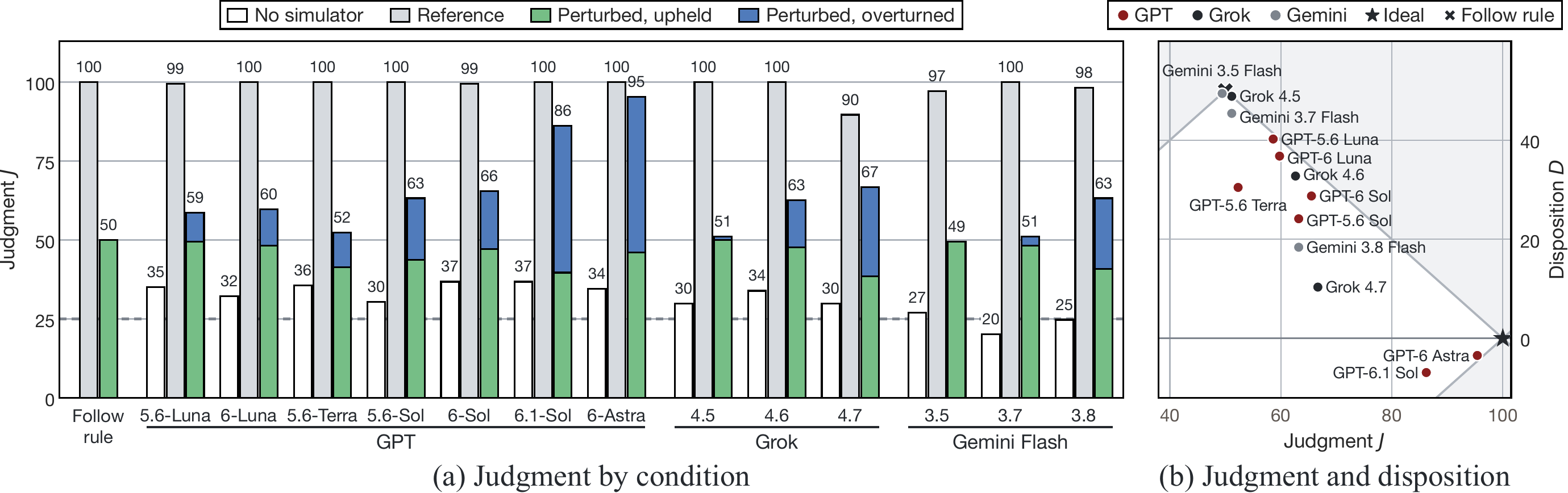}
\caption{Final choices under each evidence condition. (a) Judgment, with perturbed bars split by problem type and random choice dashed. (b) Judgment and disposition under the perturbed simulator, unreachable regions shaded.}
\label{fig:conditions}
\end{figure*}

\niparagraph{Simulator and candidate designs.}
We instantiate \sysname for LLM serving with a simulator based on LLMServingSim~\cite{llmservingsim:iiswc:2024}\footnote{\url{https://github.com/casys-kaist/LLMServingSim}}, whose unperturbed version serves as the reference.
The simulator's cost parameters are per-step costs measured on GPUs and interpolated over token counts, and its configuration fixes the GPUs, model, replicas, and KV cache memory.
The four candidates, shown under neutral labels, port designs from recent serving systems for three design problems: request routing~\cite{sglang:neurips:2024,preble:iclr:2025,dualmap:iclr:2026}, prefill-to-decode handoff~\cite{distserve:osdi:2024,splitwise:isca:2024}, and local scheduling with chunked prefill~\cite{sarathi-serve:osdi:2024}.
Routing and handoff score p95 end-to-end latency, and local scheduling scores p95 time to first token.
\Cref{sec:problem-kinds} gives details.

\niparagraph{Perturbations and evaluation set.}
We inject seven perturbations that omit, simplify, or miscompute parts of the simulated work, costs, or resource interactions in these design problems, such as limiting prefix-reuse credit, as controlled changes for this study rather than reported defects of LLMServingSim.
Construction and validation yield 58 problems in 29 perturbation pairs, and \Cref{tab:dataset-stats} in the appendix breaks them down by problem kind.

\niparagraph{Models and repetitions.}
We evaluate thirteen GPT, Grok, and Gemini LLMs as agents with high reasoning effort, repeating each problem three times per condition and pooling the repetitions.

\niparagraph{Condition coverage.}
All thirteen models run every problem in the no-simulator, perturbed, and reference conditions, and GPT-6.1 Sol and GPT-5.6 Sol also run three evidence-component conditions.
These keep the perturbed simulator's code, cost parameters, and configuration and add neither, only the execution records, or only the predictions.

\section{Results}
\label{sec:results}
\label{sec:mechanism-diagnostics}

\subsection{Simulator Evidence}
\label{sec:condition-accuracy}
\niparagraph{Judgment by condition.}
\Cref{fig:conditions}(a) reports each model's judgment on its final choices under each evidence condition.
As a baseline, the follow rule always selects the presented recommendation.
It solves every upheld problem and no overturned problem, scoring $J=50$ and $D=+50$ under the perturbed simulator.
Without simulator evidence, every model scores near chance, and with the reference simulator, all but Grok 4.7 reach near-perfect judgment.
Neither condition thus reveals how well a model judges simulator evidence.

\niparagraph{Perturbed simulator.}
The perturbed bars in \cref{fig:conditions}(a) show that every model's judgment rises above its no-simulator level.
This improvement tends to grow with model recency within each family.
GPT-6 Astra and GPT-6.1 Sol approach their reference-simulator judgment, whereas the other models stay closer to the follow rule.
Perturbed simulator evidence thus remains useful, and judgment under it separates models that accuracy without a simulator or with a sound one cannot distinguish, making it more informative for ADRS.
\Cref{tab:conditions} in the appendix gives exact values.

\niparagraph{Distinct decision profiles.}
\Cref{fig:conditions}(b) places each model's final choices under the perturbed simulator by judgment and disposition.
Judgment grows to the right, and positive disposition means higher accuracy on upheld than on overturned problems.
GPT-6 Astra and GPT-6.1 Sol occupy the region with high judgment and similar upheld and overturned accuracies.
Models with similar judgment can nevertheless differ substantially in disposition.
GPT-6 Sol and Grok 4.7, for example, score 66 and 67 in judgment but $+29$ and $+10$ in disposition.
The two panels together thus show that perturbed simulator evidence distinguishes the models in both overall accuracy and decision behavior.

\begin{figure}[t]
\centering
\includegraphics[width=\columnwidth]{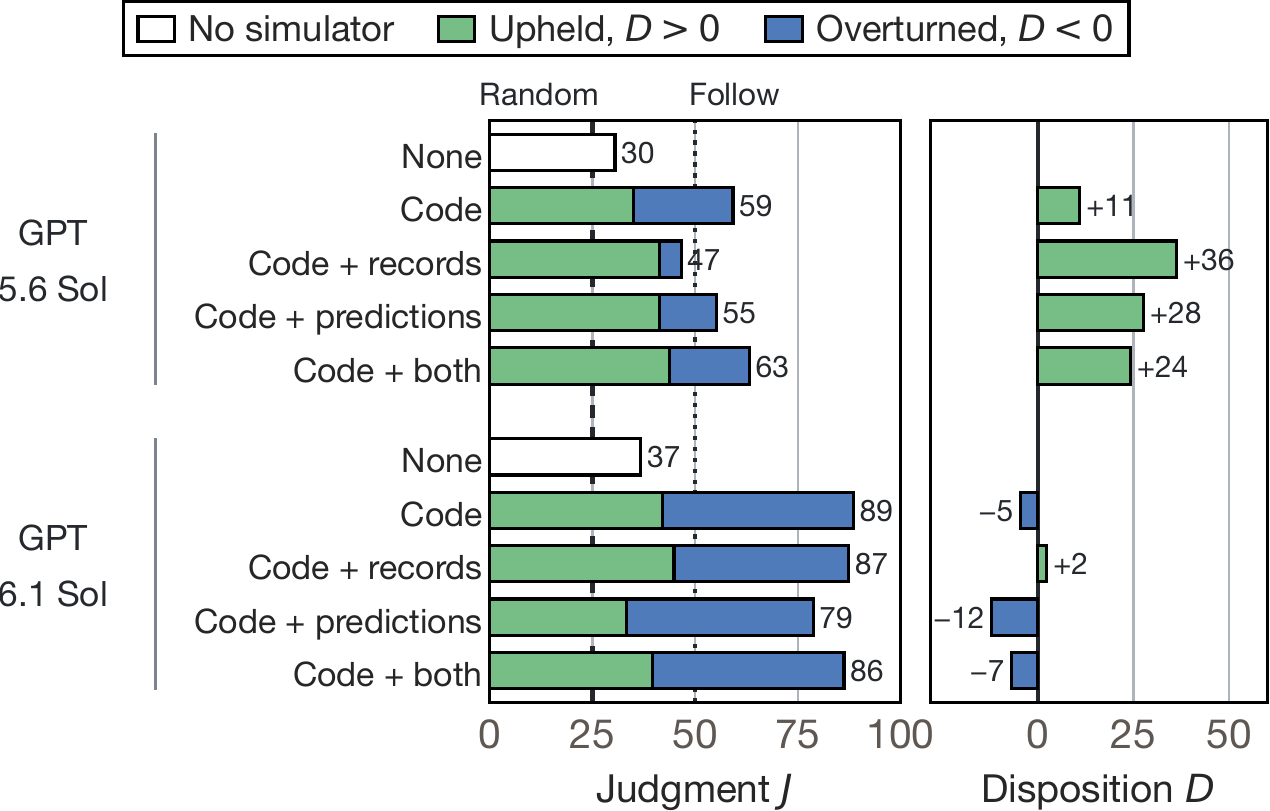}
\caption{Judgment and disposition by the simulator evidence available. Code denotes the simulator code, cost parameters, and configuration.}
\label{fig:evidence-components}
\end{figure}

\niparagraph{Evidence components.}
We next vary the evidence components for GPT-5.6 Sol and GPT-6.1 Sol, which have distinct dispositions in \cref{fig:conditions}(b).
Each model receives the simulator code, cost parameters, and configuration, either alone or with one of the simulator's outputs, the execution records or the predictions.
\Cref{fig:evidence-components}, which labels this bundle Code, shows that Code alone yields judgment close to or above that with the full perturbed evidence, although the simulator code contains the perturbation.
Adding either output lowers judgment in both models, but their effects on disposition differ.
Execution records raise disposition in both models, whereas predictions raise it for GPT-5.6 Sol and lower it for GPT-6.1 Sol.
For these two models, most of the gain from simulator evidence is thus available without its outputs, and the outputs shift the balance between problem types without consistently improving judgment.
\Cref{tab:evidence-split} in the appendix gives exact values for both models.

\subsection{Decision Impact}
\label{sec:selection-performance}
\label{sec:identification-selection}
\niparagraph{Paired decisions.}
\Cref{fig:conditions}(b) separates models that perform well on both problem types from those whose correct choices concentrate on upheld problems.
Solving both problems of a perturbation pair requires keeping the presented recommendation on one workload and replacing it on the other, even though both problems contain the same limitation.
\Cref{tab:effect-judgments} in the appendix shows that GPT-6 Astra and GPT-6.1 Sol solve both problems in 79 and 65 of 87 pair attempts, 29 pairs in three repetitions, whereas every other model solves fewer than half.
We next relate these choices to limitation identification.

\begin{figure}[t]
\centering
\includegraphics[width=\columnwidth]{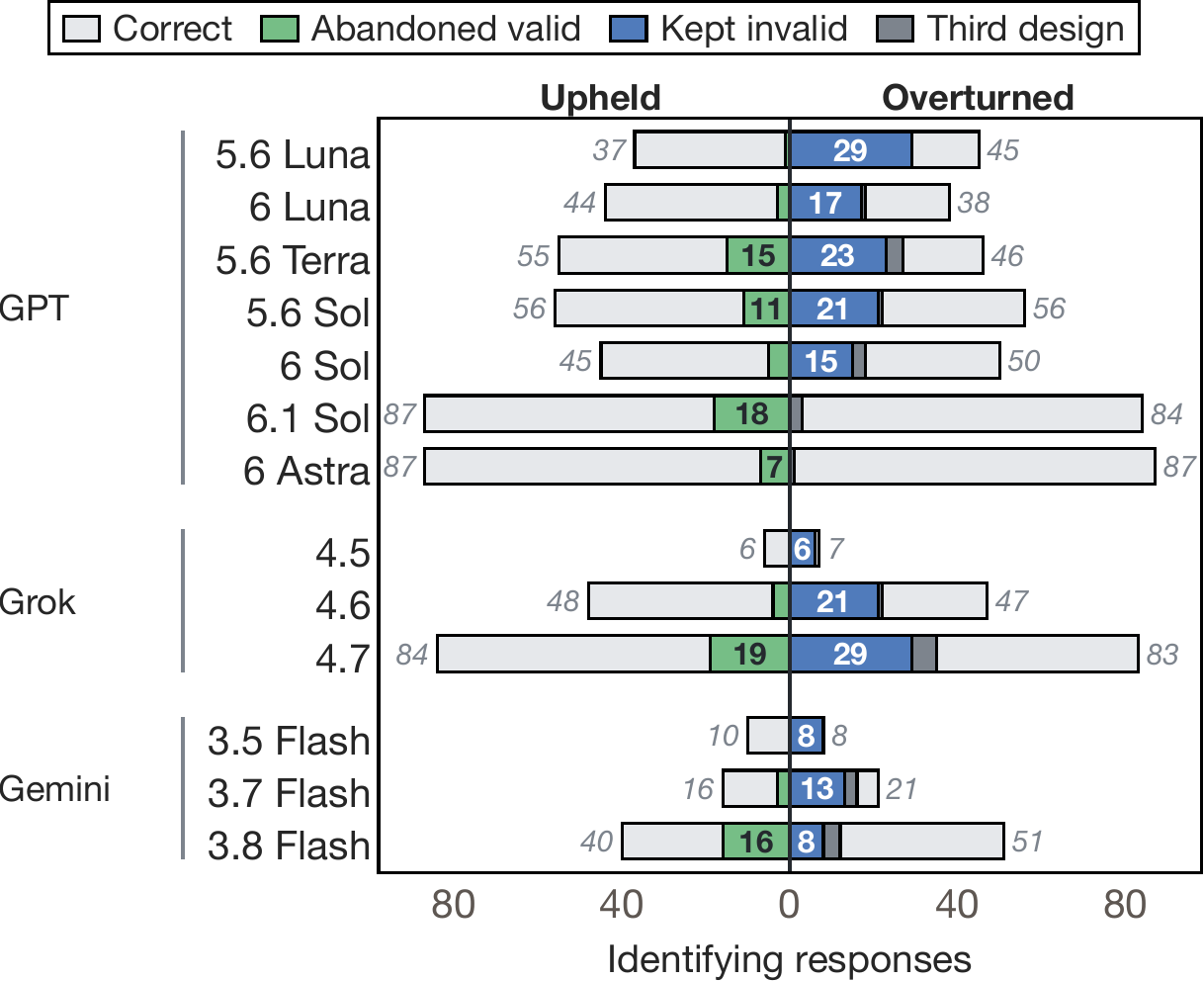}
\caption{Decisions among responses that identify the limitation. End labels count identifying responses.}
\label{fig:identification}
\end{figure}

\niparagraph{Decisions after identification.}
\Cref{fig:identification} shows that about a quarter of the responses that identify the limitation still select a design other than the reference answer, and every model makes such errors.
Even GPT-6 Astra and GPT-6.1 Sol, which identify it in nearly every response, still abandon some valid recommendations, on upheld problems where the limitation changes predicted values but leaves the reference answer unchanged.
These errors increase p95 latency under the reference simulator by a median of 14\% per problem for the eight models with loss measurements, as \cref{sec:result-counts} details.
Identifying the limitation therefore does not ensure a correct decision, and evaluating identification alone would miss these errors.
\Cref{tab:failure-types} in the appendix gives the exact counts for each error type.

\begin{figure}[t]
\centering
\includegraphics[width=\columnwidth]{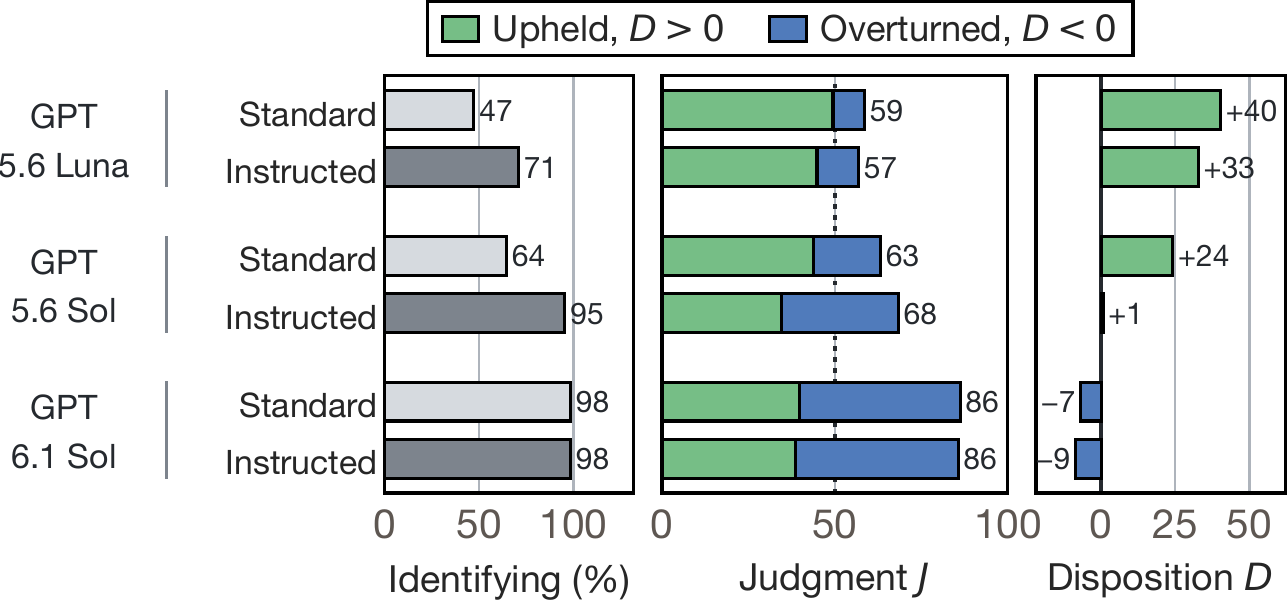}
\caption{Perturbed-simulator results with the standard and instructed-inspection prompts. Identifying is the share of responses that identify the limitation.}
\label{fig:instructed}
\end{figure}

\niparagraph{Instructed inspection.}
The share of conversations that read the simulator code changed by the perturbation ranges from 13\% to 100\% across models, with similar shares under both simulators.
Several models read it mostly during the review.
GPT-5.6 Sol and GPT-5.6 Luna, for example, read it in 80\% and 72\% of conversations but in only 4\% and 2\% of first turns.
\Cref{tab:inspection} in the appendix gives the rates per model, turn, and simulator.

\Cref{fig:instructed} therefore compares GPT-6.1 Sol, GPT-5.6 Sol, and GPT-5.6 Luna under the perturbed simulator with and without an instruction to read this code and check it against the system specification before recommending.
The instruction substantially increases limitation identification in GPT-5.6 Sol and GPT-5.6 Luna, but does not consistently improve their judgment.
Their disposition changes more than their judgment, while GPT-6.1 Sol, which already inspects this code, changes little.
Because the instruction also suggests a mismatch between the simulator and the specification, this comparison does not isolate the effect of inspection.
Increased limitation identification thus does not reliably improve decisions on whether a recommendation should stand.
\Cref{tab:instructed-inspection} in the appendix gives exact counts for all three models.

\subsection{Self-Correction}
\label{sec:self-correction}
\begin{figure}[t]
\centering
\includegraphics[width=\columnwidth]{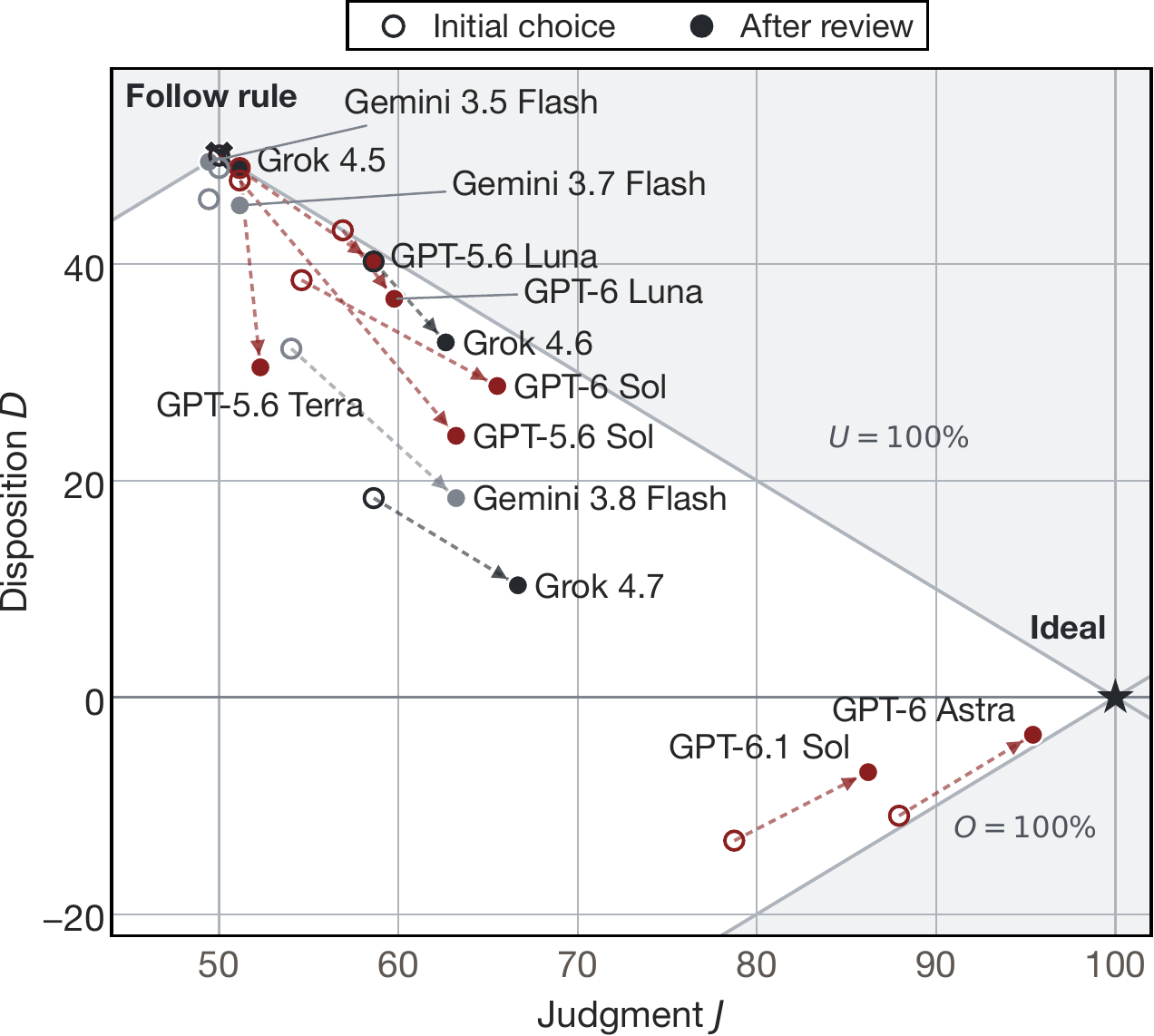}
\caption{Change from initial (hollow) to final (filled) choices through the review turn.}
\label{fig:selection}
\end{figure}

\niparagraph{Review trajectories.}
\Cref{fig:selection} connects each model's initial choices under the perturbed simulator to the final choices in \cref{fig:conditions}(b), so each arrow's horizontal and vertical extents give the review's $\Delta J$ and $\Delta D$.
GPT-6 Astra and GPT-6.1 Sol begin with high overturned accuracy and improve mainly by restoring valid recommendations abandoned in the first turn.
Several other models correct invalid recommendations they initially kept, while three remain near the follow rule.
Overall, twelve of the thirteen models move toward more similar upheld and overturned accuracies.

\niparagraph{Corrections and new errors.}
These movements yield different gains in judgment.
\Cref{tab:review-movement} in the appendix shows that review corrects 31 choices and introduces 10 new errors for GPT-5.6 Sol, raising $J$ by 12 points.
For GPT-5.6 Terra, 17 corrections are offset by 15 new errors, yielding only a 1-point gain despite a large change in disposition.
Review thus tends to reduce the imbalance between problem types, while its gains in judgment vary across models and can be offset by new errors.

\section{Discussion}
\label{sec:discussion}

\niparagraph{Broadening the problem space.}
While our case study focuses on LLM serving, \sysname can extend to other system domains whose simulators likewise trade fidelity for speed and scale in ways that can change design decisions, as \Cref{tab:abstractions} shows.
Doing so requires a reference that identifies the best design per workload, workloads on which a perturbation yields both upheld and overturned problems, and inspectable evidence.
Documented limitations of existing simulators could further supply naturally occurring problems when a reference establishes their decision impact.
Future work could also inject several limitations at once, whose effects may compound or cancel out, and define decision impact by their combined effect.
It could likewise move beyond multiple-choice selection of the best design to other question formats, such as ranking all candidates or proposing a revised design, defining decision impact for each format.

\niparagraph{Integrating \sysname into ADRS.}
Beyond design review task, \sysname can be extended to iterative ADRS workflows to study how judgments shape later proposals and final designs.
In workflows that select candidates directly by evaluator scores, such as AlphaEvolve~\cite{alphaevolve:arxiv:2025}, search pressure may favor designs that exploit simulator limitations, so an extension could test whether an explicit LLM review results improves the search.
In workflows where LLMs already interpret simulation results, such as Glia~\cite{glia:cais:2026}, the evaluation could track whether they retain or override the simulator's recommendation across iterations.
When LLMs also modify the simulator, as in Database ADRS~\cite{adrs-databases:arxiv:2026}, the evaluation would need to record the designs and evidence at each iteration, and to keep a reference independent of the evolving simulator.
This would connect judgment on individual decisions to the designs that ADRS ultimately produces.

\niparagraph{Choosing additional evidence.}
\sysname could also evaluate whether LLMs decide what to measure before selecting a design.
When the presented evidence leaves a comparison unresolved, an LLM could request a targeted measurement under a budget, testing whether it recognizes the uncertainty and gathers information that resolves it.
Since more evidence does not uniformly improve decisions in our study, scoring should weigh the final design against the cost of the measurements.

\section{Related Work}
\label{sec:related-work}

\niparagraph{Simulator validation.}
Simulation validation assesses fidelity relative to intended use~\cite{network-simulation-validation:ieee-network:2001}, and Proteus~\cite{proteus:tpds:2024} also checks whether its predictions preserve the ordering of parallelization strategies.
Simthesizer~\cite{simthesizer:arxiv:2026} has a coding agent extend a serving simulator from feature requests and validates extensions by throughput error.
Decision-focused learning evaluates predictions by the quality of decisions they induce~\cite{decision-focused:aaai:2019,smart-predict-optimize:mansci:2022}, and \sysname evaluates whether agents judge this decision impact when reading a simulator with a modeling limitation.

\niparagraph{Research agents and review.}
Research-agent benchmarks such as MLE-bench~\cite{mle-bench:iclr:2025}, RE-Bench~\cite{re-bench:icml:2025}, and ScienceAgentBench~\cite{scienceagentbench:iclr:2025} score completed research tasks.
MIRAGE~\cite{mirage:eacl:2026} varies guidance fidelity and contextual relevance, while \citet{knowledge-conflicts:iclr:2024} study conflicts between external evidence and parametric knowledge.
\citet{self-correct-reasoning:iclr:2024} find that intrinsic self-correction often fails on reasoning tasks, and \citet{self-correction-survey:tacl:2024} conclude that it helps mainly when reliable external feedback exists.
In code review, prompts asking for explanations and fixes can increase rejection of correct implementations~\cite{code-review-overcorrection:ase:2026}.
\sysname scores one design decision under controlled simulator evidence and records the corrections and new errors of a review turn without added feedback.

\section{Conclusion}
\label{sec:conclusion}

\sysname evaluates whether LLMs judge the decision impact of simulator modeling limitations, using workload pairs on which the same limitation upholds or overturns the presented recommendation.
Across thirteen LLMs, perturbed simulator evidence reveals differences in design decisions that accurate recommendations largely obscure.
Identifying a limitation does not ensure a correct choice.
Inspection instructions and self-correction often shift disposition more than they raise judgment.
Evaluating LLMs for ADRS therefore requires scoring the decisions that follow limitation identification, counting both error directions together.
We plan to publicly release the \sysname artifact.

\section*{Limitations}

\niparagraph{Measured construct.}
\sysname measures systems understanding through design choices under controlled simulator limitations.
High accuracy shows successful decisions on these problems but does not establish which reasoning produced each choice or how it transfers to a complete ADRS workflow.

\niparagraph{Reference-defined answers.}
Correct choices are defined by the reference simulator under a fixed system specification, and the reference is not a perfect model of the real system.
Upheld and overturned labels concern the top-1 design, while smaller margins and lower-rank changes can also matter to a design decision.
Selection loss likewise measures a gap within the reference simulator.

\niparagraph{Real-system outcomes.}
A choice that matches the reference answer is correct for a particular specification and simulator, and its advantage on real hardware is untested.
The real-system illustration in \Cref{sec:real-system-cases} covers particular policies, one workload trace, two A100 GPUs, and two launches.
Its p95 TTFT winner is a policy that both simulator versions recommend, and its throughput winner is unresolved, so it shows that a simulator comparison can connect to measured performance and contains no case in which an LLM improves real performance by overturning an invalid recommendation.

\niparagraph{Problem space and exposure.}
The study covers one domain, LLM serving, one simulator family, and one perturbation per problem.
The 29 pairs cover three design problems and seven perturbations selected for review, and each change form appears in one design problem, so the study cannot separate the effect of the problem from that of the perturbation.
All problems were constructed and inspected during development, and repetitions vary responses on the same problems without adding new situations.
The results compare the tested models on these problems and do not establish a general model ranking or deployment performance.

\niparagraph{Evidence components.}
The evidence-component study covers two models and provides the simulator code, cost parameters, and configuration together.
It therefore characterizes this evidence bundle and the effect of adding simulator outputs, without separating the contribution of the code from that of the numerical system information.

\niparagraph{Identification annotation.}
Limitation identification is annotated from public responses by LLM annotators that agree closely, which shows reproducible labels but cannot remove biases they share.
A stated limitation records what the LLM wrote, and the review asks only for consequential limitations, so neither statement nor omission reveals the computation behind a choice.

\niparagraph{Self-correction through review.}
We call changes between the two turns self-correction, but a single review turn is our only probe of it, so the results describe the effect of this turn on design choices.
Besides the same evidence packet, the review lets the LLM open files it skipped and directs attention to the simulator and its modeling limitations.
Its corrections and new errors therefore combine reconsideration with extra computation, newly read files, and prompt-induced doubt.
Separating these factors requires a continuation with the same extra computation or an independent second attempt, which we did not run, so the results do not show how well these models self-correct in general or whether review outperforms other uses of extra inference~\cite{self-correction-survey:tacl:2024}.

\niparagraph{Inference settings.}
Reasoning effort is the requested setting, which does not imply equal computation across providers.

\section*{Acknowledgments}

This research was supported by the MSIT(Ministry of Science, ICT), Korea, under the Global Research Support Program in the Digital Field program(RS-2024-00436680) supervised by the IITP(Institute for Information \& Communications Technology Planning \& Evaluation).
This project is supported by Microsoft Research Asia.
We used Copilot in this research.

\bibliography{paper}

\clearpage
\appendix
\crefalias{section}{appendix}
\crefalias{subsection}{appendix}
\section{Protocol and Collection Details}
\label{sec:protocol-details}

\subsection{Construction and Coverage}
\label{sec:construction-coverage}
\begin{table}[!b]
\centering
\small
\setlength{\tabcolsep}{4pt}
\begin{tabular}{@{}lrrrr@{}}
\toprule
 & & & \multicolumn{2}{c}{Conversations} \\
\cmidrule(l){4-5}
Problem kind & Pairs & Problems & Per cond. & Total \\
\midrule
\textit{Request routing} & \textit{12} & \textit{24} & \textit{936} & \textit{2,808} \\
\quad Prefix cap & 6 & 12 & 468 & 1,404 \\
\quad Reuse granularity & 6 & 12 & 468 & 1,404 \\
\midrule
\textit{Handoff} & \textit{6} & \textit{12} & \textit{468} & \textit{1,404} \\
\quad Shared NIC & 6 & 12 & 468 & 1,404 \\
\midrule
\textit{Local scheduling} & \textit{11} & \textit{22} & \textit{858} & \textit{2,574} \\
\quad Query interp. & 5 & 10 & 390 & 1,170 \\
\quad Step aggregation & 2 & 4 & 156 & 468 \\
\quad Coordinate swap & 2 & 4 & 156 & 468 \\
\quad Interp. arithmetic & 2 & 4 & 156 & 468 \\
\midrule
\textbf{Total} & \textbf{29} & \textbf{58} & \textbf{2,262} & \textbf{6,786} \\
\bottomrule
\end{tabular}
\caption{Evaluation set by problem kind. Each pair has one upheld and one overturned problem, and conversations count those assigned to thirteen models with three repetitions in each of the three evidence conditions.}
\label{tab:dataset-stats}
\end{table}

\niparagraph{Perturbation scope.}
Each perturbation realizes an abstraction error or an incorrect implementation of a specified computation, two sources of simulation error~\cite{simulation-error:ispass:2014} that a reader can detect by comparing the simulator source with the specification.
The evaluated changes therefore form a selected sample of simulator errors, and uncertainty in the specification, data validity~\cite{simulation-vv:jos:2013}, and interacting limitations lie outside this evaluation.

\niparagraph{Admission criteria.}
The current evaluation set admits a problem when its execution is valid and repeatable, both simulator versions have a unique winner, the changed computation executes and alters predicted values in upheld problems, and the presented evidence supports the reference answer.
\Cref{sec:evidence-validation} describes the evidence-validation procedure and its exclusions.

\niparagraph{Adaptive proposals.}
An LLM Builder proposes policies and trace generators within fixed construction criteria and revises proposals using simulator feedback.
Every executed proposal freezes its code, exact input, sources, and measurement contract.
An independent Critic and Arbiter review the evidence, while automatic failures remain binding.
If confirmation informed a revision, the changed proposal requires fresh confirmation.

\niparagraph{Construction roles.}
The discovery and confirmation traces define simulator-construction roles independently of agent-evaluation splits.
Development exposure, earlier failed proposals, and exact-trace-only qualification retain their original status.
Metrics sharing a simulator execution do not multiply independent workload samples.

\subsection{Data Composition}
\label{sec:data-composition}
\niparagraph{Reuse of earlier material.}
Candidates, workloads, and simulator executions from an earlier problem bank, which contains only overturned problems, enter the evaluation only under the same eligibility, pairing, and public-evidence criteria as new material.
These criteria include the packet validation in \Cref{sec:evidence-validation}, and the evaluation set was fixed before the evaluated agents responded.
The construction records document reused items, exclusion reasons, and development exposure.

\subsection{Real-System Case Protocol}
\niparagraph{Policies and input.}
The illustrative measurement uses 144 requests, arriving eight at a time every 0.12 seconds.
Input lengths are 64, 256, 768, and 1,536 tokens, and output lengths are 16, 64, and 96 tokens, with 12 requests per length combination.
All four policies process the same trace, with 32 total measured executions across two launches.
Warm-up and environment checks are excluded from the reported metrics.

\niparagraph{Uncertainty procedure.}
The recorded procedure jointly resamples policy vectors within each launch using 20,000 bootstrap draws.
For each metric, it constructs simultaneous 95\% intervals over the six policy pairs.
A resolved winner must beat all three alternatives under these intervals, rank uniquely first in at least seven of eight repetitions, and lead within each launch and overall.
The intervals are conditional on the observed launches and have limited tail-probability support from four repetitions per launch.

\subsection{Evidence Replay}
\niparagraph{Replay checks.}
The replay protocol reconstructs question assembly, retrieval, response scoring, and aggregation from fixed evidence.
Negative checks test mismatched sources, labels, instructions, and public-evidence boundaries.
Fresh-environment checks compare the resulting questions, field scores, and analysis denominators with their stored references.

\niparagraph{Reuse contract.}
The protocol fixes problem identity, condition matching, activation and recommendation checks, and the connection from presented evidence to choices.
A different environment must supply policy, workload, metric, and simulator semantics together with inspectable evidence.
Evaluating another agent on fixed evidence can reuse the same simulator records.
The serving implementation supplies these bindings for scheduling policies and simulator costs.

\section{Design-Review Protocol}
\label{sec:mechanism-diagnostic-protocol}

\subsection{Serving Problem Kinds}
\label{sec:problem-kinds}
\begin{table*}[t]
\centering
\small
\begin{tabular}{@{}p{3.3cm}p{8.6cm}p{3.2cm}@{}}
\toprule
Problem kind & Perturbed computation & Change form \\
\midrule
\multicolumn{3}{@{}l}{\textbf{Request routing}: which replica serves each request, among load-only, SGLang-inspired, Preble, and DualMap} \\
\quad Prefix cap & Prefix reuse is credited only up to 512 tokens & Value or constant \\
\quad Reuse granularity & Prefix reuse is credited only in 256-token units & Value or constant \\
\midrule
\multicolumn{3}{@{}l}{\textbf{Prefill-to-decode handoff}: where decode runs after prefill, among stay on source and three remote placements} \\
\quad Shared NIC & Transfers use a dedicated link per GPU pair instead of a shared NIC & Omission \\
\midrule
\multicolumn{3}{@{}l}{\textbf{Local scheduling}: the prefill chunk cap per step, among 16, 24, and 32 tokens and no cap} \\
\quad Query interpolation & Step-cost lookup omits interpolation along the query axis & Omission \\
\quad Step aggregation & Step cost comes from the largest request instead of the sum & Aggregation \\
\quad Coordinate swap & Step-cost lookup swaps the query and KV coordinates & Variable or coordinate \\
\quad Interpolation arithmetic & Interpolation computes $y_0+r\,y_1$ instead of $y_0+r(y_1-y_0)$ & Arithmetic or sign \\
\bottomrule
\end{tabular}
\caption{Design problems and problem kinds in the serving instantiation. Each design problem compares four designs, and each problem kind perturbs one simulator computation that all four rely on.}
\label{tab:families}
\end{table*}

\niparagraph{Change forms.}
The seven perturbations cover five forms adapted from mutation operators: replacing a value, omitting a cost or contention term, changing how costs aggregate, swapping lookup coordinates, and altering interpolation arithmetic.
They are controlled changes for this study rather than reported defects of LLMServingSim.

\niparagraph{Routing.}
Load-only, SGLang-inspired, Preble, and DualMap send the same requests to different replicas, so the same prefix-accounting change produces different executions and costs for each.
Workloads vary shared prefixes and load.

\niparagraph{Handoff and local scheduling.}
Handoff compares staying on the prefill source with three remote decode placements, and document length sets the size of the state each transfer moves.
Local scheduling compares prefill chunk caps of 16, 24, 32, and full, and output length changes concurrent decode and therefore which cost lookups the perturbation affects.

\begin{table}[t]
\centering
\small
\setlength{\tabcolsep}{2.6pt}
\begin{tabular}{@{}lcccccccc@{}}
\toprule
 & \multicolumn{4}{c}{Upheld} & \multicolumn{4}{c}{Overturned} \\
\cmidrule(lr){2-5}\cmidrule(l){6-9}
Model & Init. & $+$ & $-$ & Fin. & Init. & $+$ & $-$ & Fin. \\
\midrule
GPT-6 Astra & 67 & 14 & 1 & 80 & 86 & 0 & 0 & 86 \\
GPT-6.1 Sol & 57 & 13 & 1 & 69 & 80 & 1 & 0 & 81 \\
GPT-6 Sol & 81 & 3 & 2 & 82 & 14 & 18 & 0 & 32 \\
GPT-5.6 Sol & 86 & 0 & 10 & 76 & 3 & 31 & 0 & 34 \\
GPT-5.6 Terra & 87 & 0 & 15 & 72 & 2 & 17 & 0 & 19 \\
GPT-6 Luna & 87 & 0 & 3 & 84 & 12 & 8 & 0 & 20 \\
GPT-5.6 Luna & 87 & 0 & 1 & 86 & 0 & 16 & 0 & 16 \\
Grok 4.5 & 87 & 0 & 0 & 87 & 0 & 2 & 0 & 2 \\
Grok 4.6 & 86 & 0 & 3 & 83 & 16 & 10 & 0 & 26 \\
Grok 4.7 & 67 & 6 & 6 & 67 & 35 & 17 & 3 & 49 \\
Gemini 3.5 Flash & 86 & 0 & 0 & 86 & 1 & 0 & 1 & 0 \\
Gemini 3.7 Flash & 83 & 1 & 0 & 84 & 3 & 2 & 0 & 5 \\
Gemini 3.8 Flash & 75 & 0 & 4 & 71 & 19 & 20 & 0 & 39 \\
\bottomrule
\end{tabular}
\caption{Initial-to-final movement of perturbed choices. Init.\ and Fin.\ count correct initial and final choices, and $+$ and $-$ count choices made correct and incorrect in review, so $+$ is bounded by the 87 minus Init.\ initially wrong choices and $-$ by the Init.\ initially correct ones. Init.\ and Fin.\ give $J$ and $D$ before and after review through \cref{eq:judgment-disposition}.}
\label{tab:review-movement}
\end{table}

\begin{table}[t]
\centering
\small
\setlength{\tabcolsep}{4.5pt}
\begin{tabular}{@{}lcccccc@{}}
\toprule
 & No-sim & Ref. & \multicolumn{4}{c}{Perturbed} \\
\cmidrule(lr){2-2}\cmidrule(lr){3-3}\cmidrule(l){4-7}
Model & $J$ & $J$ & Up. & Ov. & $J$ & $D$ \\
\midrule
GPT-6 Astra & 34 & 100 & 92 & 99 & 95 & $-3$ \\
GPT-6.1 Sol & 37 & 100 & 79 & 93 & 86 & $-7$ \\
GPT-6 Sol & 37 & 99 & 94 & 37 & 66 & $+29$ \\
GPT-5.6 Sol & 30 & 100 & 87 & 39 & 63 & $+24$ \\
GPT-5.6 Terra & 36 & 100 & 83 & 22 & 52 & $+30$ \\
GPT-6 Luna & 32 & 100 & 97 & 23 & 60 & $+37$ \\
GPT-5.6 Luna & 35 & 99 & 99 & 18 & 59 & $+40$ \\
Grok 4.5 & 30 & 100 & 100 & 2 & 51 & $+49$ \\
Grok 4.6 & 34 & 100 & 95 & 30 & 63 & $+33$ \\
Grok 4.7 & 30 & 90 & 77 & 56 & 67 & $+10$ \\
Gemini 3.5 Flash & 27 & 97 & 99 & 0 & 49 & $+49$ \\
Gemini 3.7 Flash & 20 & 100 & 97 & 6 & 51 & $+45$ \\
Gemini 3.8 Flash & 25 & 98 & 82 & 45 & 63 & $+18$ \\
\midrule
Follow rule & -- & 100 & 100 & 0 & 50 & $+50$ \\
\bottomrule
\end{tabular}
\caption{Judgment $J$, disposition $D$, and upheld (Up.) and overturned (Ov.) accuracy (\%) by condition.}
\label{tab:conditions}
\end{table}

\begin{table}[t]
\centering
\small
\setlength{\tabcolsep}{4.5pt}
\begin{tabular}{@{}lcccccc@{}}
\toprule
Evidence & Up. & Ov. & $J$ & $D$ & Pair & Follow \\
\midrule
\multicolumn{7}{@{}l}{\textbf{GPT-5.6 Sol}} \\
None & 39 & 22 & 30 & $+9$ & 10 & -- \\
Code & 70 & 48 & 59 & $+11$ & 37 & 31 \\
Code + records & 83 & 10 & 47 & $+36$ & 8 & 69 \\
Code + predictions & 83 & 28 & 55 & $+28$ & 22 & 59 \\
Code + both & 87 & 39 & 63 & $+24$ & 34 & -- \\
Code, correct & 67 & 70 & 68 & $-2$ & 46 & -- \\
\midrule
\multicolumn{7}{@{}l}{\textbf{GPT-6.1 Sol}} \\
None & 29 & 45 & 37 & $-8$ & 9 & -- \\
Code & 84 & 93 & 89 & $-5$ & 77 & 3 \\
Code + records & 90 & 85 & 87 & $+2$ & 76 & 10 \\
Code + predictions & 67 & 91 & 79 & $-12$ & 62 & 3 \\
Code + both & 79 & 93 & 86 & $-7$ & 75 & -- \\
Code, correct & 89 & 98 & 93 & $-5$ & 87 & -- \\
\bottomrule
\end{tabular}
\caption{Results by the simulator evidence available. Up., Ov., and Pair are in percent. Code is the simulator code, cost parameters, and configuration, each Code + row adds evidence to Code alone, Code + both is the full perturbed evidence, and Code, correct gives the code of the reference simulator. Follow counts overturned problems, of 87, on which the final choice is the perturbed simulator's top design, shown or not.}
\label{tab:evidence-split}
\end{table}

\begin{table}[t]
\centering
\small
\setlength{\tabcolsep}{4.5pt}
\begin{tabular}{@{}lcccc@{}}
\toprule
 & \multicolumn{2}{c}{Any turn} & \multicolumn{2}{c}{First turn} \\
\cmidrule(lr){2-3}\cmidrule(l){4-5}
Model & Pert. & Ref. & Pert. & Ref. \\
\midrule
GPT-6 Astra & 100 & 99 & 100 & 97 \\
GPT-6.1 Sol & 100 & 100 & 100 & 97 \\
GPT-6 Sol & 74 & 76 & 34 & 38 \\
GPT-5.6 Sol & 80 & 80 & 4 & 0 \\
GPT-5.6 Terra & 82 & 74 & 13 & 6 \\
GPT-6 Luna & 68 & 65 & 27 & 27 \\
GPT-5.6 Luna & 72 & 64 & 2 & 3 \\
Grok 4.5 & 13 & 13 & 0 & 0 \\
Grok 4.6 & 61 & 65 & 32 & 28 \\
Grok 4.7 & 99 & 98 & 92 & 93 \\
Gemini 3.5 Flash & 18 & 18 & 2 & 2 \\
Gemini 3.7 Flash & 45 & 49 & 39 & 40 \\
Gemini 3.8 Flash & 74 & 81 & 63 & 69 \\
\bottomrule
\end{tabular}
\caption{Share of conversations that reach the simulator code changed by the perturbation, in percent, under the perturbed and reference simulators.}
\label{tab:inspection}
\end{table}

\begin{table}[t]
\centering
\small
\setlength{\tabcolsep}{4pt}
\begin{tabular}{@{}llccccc@{}}
\toprule
Model & Prompt & Reach & Ident. & Up. & Ov. & Pair \\
\midrule
GPT-6.1 Sol & Standard & 174 & 171 & 69 & 81 & 65 \\
 & Instructed & 174 & 171 & 67 & 82 & 65 \\
\midrule
GPT-5.6 Sol & Standard & 7 & 112 & 76 & 34 & 30 \\
 & Instructed & 173 & 166 & 60 & 59 & 40 \\
\midrule
GPT-5.6 Luna & Standard & 4 & 82 & 86 & 16 & 16 \\
 & Instructed & 162 & 123 & 78 & 21 & 18 \\
\bottomrule
\end{tabular}
\caption{Perturbed results with standard and instructed-inspection prompts. Reach counts first-turn conversations reaching the changed code, and Ident.\ counts responses identifying the limitation, both of 174. Up., Ov., and Pair count solved problems and pairs of 87.}
\label{tab:instructed-inspection}
\end{table}

\subsection{Result Counts}
\label{sec:result-counts}
\niparagraph{Exact counts.}
\Cref{tab:conditions} gives the accuracies behind \cref{fig:conditions}, and \Cref{tab:effect-judgments,tab:failure-types,tab:interventions} give the counts behind \Cref{fig:selection,fig:identification} and the auxiliary contrasts.
Selection loss is the relative p95 gap between the selected design and the best design under the reference simulator, summarized with one median per problem so that repeated responses to the same input do not dominate.
For the eight models with loss measurements, 151 of their 162 identified but incorrect choices have a p95 at least 5\% above the best design, and median per-problem losses are 14\% for upheld errors, 24\% for keeping the presented design, and 47\% for a third design.
\Cref{tab:review-movement} separates initial and final choices.
In 41 conversations, the initial response already states the limitation yet selects a wrong design that stays wrong.
Under the reference simulator, Grok 4.7 questions the measured cost parameters on each of its 18 errors and switches away from the sound recommendation.

\niparagraph{Code inspection.}
\Cref{tab:inspection} counts a conversation as reaching the changed code when a read returns, or a search matches, a simulator line where the perturbed and reference versions differ.
The instructed-inspection condition adds one sentence to the first-turn prompt: ``Before recommending, read the simulator source code that produces prediction.json and check whether it models the serving system as described in system-description.md.''
GPT-5.6 Sol and GPT-5.6 Luna first ran on 14 pairs, and their results led us to extend the condition to the other 15 pairs and to GPT-6.1 Sol.
On those 15 pairs alone, GPT-5.6 Sol solves 35 rather than 23 of 45 overturned problems and 35 rather than 40 upheld ones, the same direction as on all 29 pairs.
Because the instruction asks the agent to check the simulator against the specification, it may also suggest that the two disagree, and part of the upheld loss may reflect this suggestion.
\Cref{tab:instructed-inspection} gives the counts.

\begin{table}[t]
\centering
\small
\setlength{\tabcolsep}{2.6pt}
\begin{tabular}{@{}lcccccc@{}}
\toprule
 & \multicolumn{4}{c}{Perturbed} & Ref. & No-sim \\
\cmidrule(lr){2-5}\cmidrule(lr){6-6}\cmidrule(l){7-7}
Model & Up. & Ov. & All\% & Pair & Pair & Pair \\
\midrule
GPT-6 Astra & 80 & 86 & 95.4 & 79 & 87 & 6 \\
GPT-6.1 Sol & 69 & 81 & 86.2 & 65 & 87 & 8 \\
GPT-6 Sol & 82 & 32 & 65.5 & 32 & 86 & 14 \\
GPT-5.6 Sol & 76 & 34 & 63.2 & 30 & 87 & 9 \\
GPT-5.6 Terra & 72 & 19 & 52.3 & 14 & 87 & 10 \\
GPT-6 Luna & 84 & 20 & 59.8 & 20 & 87 & 8 \\
GPT-5.6 Luna & 86 & 16 & 58.6 & 16 & 86 & 7 \\
Grok 4.5 & 87 & 2 & 51.1 & 2 & 87 & 1 \\
Grok 4.6 & 83 & 26 & 62.6 & 25 & 87 & 2 \\
Grok 4.7 & 67 & 49 & 66.7 & 37 & 69 & 2 \\
Gemini 3.5 Flash & 86 & 0 & 49.4 & 0 & 83 & 3 \\
Gemini 3.7 Flash & 84 & 5 & 51.1 & 5 & 87 & 1 \\
Gemini 3.8 Flash & 71 & 39 & 63.2 & 30 & 84 & 1 \\
\midrule
Follow rule & 87 & 0 & 50.0 & 0 & 87 & -- \\
\bottomrule
\end{tabular}
\caption{Solved problems and pairs per model, of 87. Up.\ and Ov.\ count upheld and overturned problems, and All is the percentage of all 174 problems solved.}
\label{tab:effect-judgments}
\end{table}

\begin{table}[t]
\centering
\small
\setlength{\tabcolsep}{2.6pt}
\begin{tabular}{@{}lrrrrrr@{}}
\toprule
 & \multicolumn{5}{c}{Identified} & Not \\
\cmidrule(lr){2-6}
Model & All & Corr. & Up. & Pres. & Third & ident. \\
\midrule
GPT-6 Astra & 174 & 166 & 7 & 0 & 1 & 0 \\
GPT-6.1 Sol & 171 & 150 & 18 & 0 & 3 & 3 \\
GPT-6 Sol & 95 & 72 & 5 & 15 & 3 & 79 \\
GPT-5.6 Sol & 112 & 79 & 11 & 21 & 1 & 62 \\
GPT-5.6 Terra & 101 & 59 & 15 & 23 & 4 & 73 \\
GPT-6 Luna & 82 & 61 & 3 & 17 & 1 & 92 \\
GPT-5.6 Luna & 82 & 52 & 1 & 29 & 0 & 92 \\
Grok 4.5 & 13 & 7 & 0 & 6 & 0 & 161 \\
Grok 4.6 & 95 & 69 & 4 & 21 & 1 & 79 \\
Grok 4.7 & 167 & 113 & 19 & 29 & 6 & 7 \\
Gemini 3.5 Flash & 18 & 10 & 0 & 8 & 0 & 154 \\
Gemini 3.7 Flash & 37 & 18 & 3 & 13 & 3 & 137 \\
Gemini 3.8 Flash & 91 & 63 & 16 & 8 & 4 & 83 \\
\bottomrule
\end{tabular}
\caption{Perturbed responses by annotated limitation identification. Up., Pres., and Third denote upheld errors and overturned errors that keep the presented design or select a third design. All and Corr.\ count identifying responses and the correct choices among them.}
\label{tab:failure-types}
\end{table}

\begin{table}[t]
\centering
\small
\setlength{\tabcolsep}{3.4pt}
\begin{tabular}{@{}lcccccc@{}}
\toprule
 & \multicolumn{3}{c}{Batch 1} & \multicolumn{3}{c}{Batch 2} \\
\cmidrule(lr){2-4}\cmidrule(l){5-7}
Condition & Up. & Ov. & Pair & Up. & Ov. & Pair \\
\midrule
Predictions only & 55 & 7 & 2 & 49 & 6 & 1 \\
Full, standard & 53 & 36 & 31 & 55 & 35 & 30 \\
Full, comparison & 59 & 38 & 37 & 60 & 37 & 37 \\
\bottomrule
\end{tabular}
\caption{Solved problems and pairs in the auxiliary conditions, out of 60 per batch.}
\label{tab:interventions}
\end{table}

\subsection{Presented Evidence and Validation}
\label{sec:evidence-validation}
\niparagraph{Conversation protocol.}
In the perturbed and reference conditions, the file list covers the system specification, candidate code, workload, simulator source, cost parameters, configuration, per-design predictions, and per-design execution records.
The no-simulator condition lists only the specification, candidate code, and workload, and its review omits the follow-or-reconsider and limitation fields.
Each turn may list, read, and search these files, and no tool executes or modifies code.
Neither turn names the limitation, its location, or the number of limitations, and the review supplies no new information, although the LLM may then read files it did not open in the first turn.

\begin{table}[t]
\centering
\small
\begin{tabular}{@{}p{0.46\columnwidth}p{0.46\columnwidth}@{}}
\toprule
Provided to the agent & Held by the evaluator \\
\midrule
System specification & Perturbation type and location \\
Candidate code, workload, target metric & Reference implementation and source difference \\
Presented simulator code, cost parameters, configuration & Reference execution, metrics, and winner \\
Presented predictions and execution records & Pair partner's packet and reference answer \\
Read-only search and reading tools & Upheld or overturned label and pair membership \\
\bottomrule
\end{tabular}
\caption{Information boundary of the design review.}
\label{tab:boundary}
\end{table}

\niparagraph{System specification.}
The specification defines placement, topology, cache-reuse semantics, and the axes, units, and interpolation of the cost parameters for each environment.
It may help an agent notice a limitation, but it states no location, fix, or reference answer.
Each specification is fixed before collection.

\niparagraph{Packet validation.}
Two blind solvers attempted each candidate problem with only the presented packet, and an evidence audit checks that this evidence supports the reference answer, using the reference execution only to confirm consistency.
Five routing candidates are excluded, either because the presented evidence cannot order the two leading designs or because the leaders fall within 1\,ms of each other when recomputed from it.

\niparagraph{Concealment.}
Answer files, the reference source, and the reference execution lie outside every tool's accessible scope.
We check prompts, returned passages, filenames, summaries, comments, and tests for target values and implementation-history cues.
All receipts match the frozen manifest of evidence trees and prompts.

\subsection{Scoring and Analysis}
\label{sec:scoring-analysis}
\niparagraph{Choices.}
A problem is solved when the final choice equals the reference answer, and pair accuracy, which the appendix tables report, requires correct final choices on both problems of a pair under the same model, condition, and repetition, the two-instance form of contrast consistency~\cite{contrast-sets:emnlp-findings:2020}.
Each problem is a separate conversation, and initial and final choices are retained separately.
Wrong choices are classified by design identity into abandoning a correct recommendation, keeping a wrong one, and replacing a wrong one with a design other than the reference answer.

\niparagraph{Rationale annotation.}
A separate LLM annotator classified each perturbed response with a fixed rubric, blind to labels, reference answers, and model identity.
The rubric records whether the response identifies the perturbed limitation, whether it did so in the first turn, whether it links the limitation to a specific design, and which impact it claims.
A second annotator independently labeled 200 responses and agreed on identification for 198 ($\kappa=0.98$).
For the eight models annotated first, 623 of the 624 responses with a valid choice that do not identify the limitation keep the presented recommendation, which follows largely from the review asking for a reason whenever the agent departs.
This annotation describes responses and never enters pair accuracy.

\niparagraph{Collection.}
Each model uses the same read-only tools under the perturbed, reference, and no-simulator conditions with three repetitions.
The main evaluation comprises 6,786 assigned conversations, thirteen models over 58 problems, three conditions, and three repetitions.
Conversations without a valid final choice remain in the assigned denominators as incorrect.

\subsection{Auxiliary Contrasts}
\label{sec:auxiliary-protocol}
\niparagraph{Arms and pairs.}
Each auxiliary batch draws one pair by a fixed seed from each of five perturbations, the two routing perturbations, handoff, query interpolation, and step aggregation, and the second batch draws only pairs unused by the first.
The predictions-only condition lists the specification, candidate code, workload, and per-design predictions, and its review asks the agent to check the recommendation against these files.
The comparison condition uses the full-evidence packet and inserts the following sentence before the review's response format: ``If you judge that a modeling limitation exists, compare whether its effect is the same for all candidates or differs between them, and whether there is evidence that overturns a candidate's existing advantage, before deciding the final candidate. Do not drop or keep the recommendation merely because a limitation exists.''
Conversation sessions cannot be forked after the first turn, so the comparison condition runs as independent conversations with the same first-turn prompt.
Each batch assigns 360 conversations, 120 per condition, and the auxiliary results are never pooled with the main evaluation.
In the first batch, the guided review restores all nine first-turn departures from valid recommendations and adds one new departure, whereas the standard review restores two of seven and adds two.
Identifying responses number 86 under the standard review and 87 under guidance, and incorrect choices among them fall from 15 to 9, while overturned problems with the presented design kept number seven and eight.
The overturned problems solved with predictions alone fall almost entirely in handoff, and in the first batch all seven repeat the same model's no-simulator choice.

\niparagraph{Representation and replay.}
Saved-evidence replay reconstructs packets, independently scores stored responses, and regenerates the reported analyses in a fresh environment.

\section{Real-System Illustration}
\label{sec:real-system-cases}

\niparagraph{Retained measurements.}
Recorded vLLM~\cite{vllm:sosp:2023} measurements connect the serving comparisons to measured performance.
One retained comparison executes Qwen3-30B-A3B~\cite{qwen3:arxiv:2025} in BF16 on two A100 40GB GPUs and compares arrival order, short-prefill priority, most-processed priority, and least-processed priority.
The latter three policies use prefill-token caps of 128, 512, and 256, respectively.
Each policy runs eight times on the same trace, with four matched repetitions in each of two fresh launches.

\niparagraph{Metric-specific outcome.}
For p95 TTFT, the short-prefill policy has a mean of 2.24 seconds across the eight run-level measurements and is uniquely best in all eight repetitions.
Its advantage over the nearest alternative has a simultaneous 95\% interval of 6.38 to 6.94 seconds under the recorded within-metric procedure.
Both simulator sources recommend that policy.
For output throughput on the same trace, the least-processed priority has the highest mean, 735.9 tokens per second, but its nearest-alternative advantage interval spans $-3.33$ to $25.69$ tokens per second, leaving the empirical winner unresolved.

\niparagraph{Role in the analysis.}
These two outcomes show how an inspectable simulator comparison connects to an actual serving decision with metric-specific uncertainty.
The resolved TTFT result supports that case, while the throughput result preserves uncertainty even with a consistent mean ordering.

\section{Case Studies}
\label{sec:review-cases}
\label{sec:case-studies}
This appendix examines two perturbation pairs from the evaluation set in detail.
The routing pair compares designs ported from recent serving systems, and the handoff pair compares decode placements after prefill.
For each pair, we describe the designs, how the perturbation affects each design's execution, and three conversations under the perturbed simulator.
The conversations show a model that identifies the limitation and keeps a valid recommendation, one that identifies it and abandons a valid recommendation, and one that corrects an invalid recommendation in the review turn.
Excerpts quote tool calls, files, and answers verbatim, abridged with ellipses and with whitespace in records condensed.
Candidate labels are those shown in each problem, so a design can carry different labels on the two workloads.

\subsection{Request Routing with Coarse Prefix Credit}
\label{sec:case-routing}

\begin{table*}[t]
\centering
\small
\begin{tabular}{@{}p{2.6cm}p{8.6cm}cc@{}}
\toprule
Design & Behavior & Upheld (s) & Overturned (s) \\
\midrule
Load-only & Sends each request to the replica with the fewest waiting and running requests, then the fewest queued tokens, and never looks at the caches. & 1.050 / 0.703 & 0.326 / 0.230 \\
SGLang-inspired & Sends each request to the replica whose cache holds the longest prefix of its prompt, breaking ties by the fewest requests and then the fewest tokens. & 1.597 / 1.233 & 0.552 / 0.180 \\
Preble & Scores each replica by its recent load, the prompt tokens it would prefill, and the cost of evicting cached prefixes weighted by recent use, and when the best reuse covers more than half of the prompt, only replicas with that reuse compete. & 3.445 / 2.737 & 0.763 / 0.660 \\
DualMap & Hashes the prompt prefix to two candidate replicas and takes the one with more cached reuse if its predicted prefill time meets a 0.6\,s TTFT target, otherwise the one with the lower predicted time. & 1.800 / 1.327 & 0.718 / 0.209 \\
\bottomrule
\end{tabular}
\caption{Routing designs and their p95 end-to-end latency under the perturbed and reference simulators. The cache-aware designs follow SGLang~\cite{sglang:neurips:2024}, Preble~\cite{preble:iclr:2025}, and DualMap~\cite{dualmap:iclr:2026}. The reference answer is load-only on the upheld workload and SGLang-inspired on the overturned one.}
\label{tab:case-routing}
\end{table*}

\niparagraph{Designs and perturbation.}
Each design decides which of four replicas, each with its own prefix cache, serves a request, and the target metric is p95 end-to-end latency.
\Cref{tab:case-routing} describes the designs, and load-only is the presented recommendation on both workloads.
When a request is first scheduled, the perturbed simulator rounds its prefix-cache hit down to a multiple of 256 tokens, whereas the system specification credits the longest cached prefix in whole 16-token blocks.
The lost credit is recomputed as prefill, and the perturbation adds these lines to the scheduler:
\begin{convtool}{Perturbation diff}
\convadd{+const PREFIX\_REUSE\_GRANULARITY\_TOKENS: u32 = 256;}
\convdel{-\ \ longest\_hit.min(hit\_ceiling)}
\convadd{+\ \ let hit = longest\_hit.min(hit\_ceiling);}
\convadd{+\ \ hit - hit \% PREFIX\_REUSE\_GRANULARITY\_TOKENS}
\end{convtool}

\niparagraph{Workloads.}
The upheld workload sends 96 measured requests 5\,ms apart over four shared 576-token prefixes with uniform popularity and 20\% of requests on one hot prefix, so load decides the ranking and load-only stays best by 75\% under the reference simulator.
The overturned workload sends 48 measured requests 22\,ms apart over eight 928-token prefixes with Zipf popularity and 40\% on one hot prefix, so reuse decides the ranking and SGLang-inspired is best by 16\%.
The rounding keeps 512 of 576 creditable tokens of an upheld hit and 768 of 928 of an overturned hit.

\niparagraph{Design-specific impact.}
The cache-affine designs hit more often and therefore lose more credit.
Over all requests of the overturned workload, the rounding raises recomputed prompt tokens by 24\% for load-only, 83\% for SGLang-inspired and Preble, and 92\% for DualMap.
The perturbed simulator therefore overestimates the latency of SGLang-inspired by 206\% and that of DualMap by 244\%, against 42\% for load-only, which reverses the ranking.
Preble loses as much credit as SGLang-inspired, but its error is 16\%, since its high latency comes from its placement.
On the upheld workload, recomputation also grows faster for SGLang-inspired than for load-only, by 161\% against 56\%, but the load advantage of load-only keeps it first.

\niparagraph{Outcomes across models.}
Of the 39 perturbed conversations per workload, 36 keep the valid upheld recommendation and 15 replace the invalid overturned one, and 12 repetitions solve both problems.
Without simulator evidence, 10 and 14 are correct.

\begin{figure}[t]
\convturn{First turn}
\begin{convuser}
\ldots\ Recommend one candidate for this workload.
\end{convuser}
\begin{convagent}{Initial choice: D}
``D has the lowest predicted measured-request p95 end-to-end latency (1.0496 s), below C (1.5966 s), A (1.7999 s), and B (3.4449 s).''
\end{convagent}
\convturn{Review turn}
\begin{convuser}
Review your recommendation before it is final. \ldots
\end{convuser}
\begin{convtool}{Tool call 8 of 21}
read \ldots/schedulers/chunked\_prefill.rs
\convadd{+const PREFIX\_REUSE\_GRANULARITY\_TOKENS: u32 = 256;}
\convadd{+ hit - hit \% PREFIX\_REUSE\_GRANULARITY\_TOKENS}
\end{convtool}
\begin{convtool}{Cited records, request 8}
D: "input\_tokens": 592, "kv": 0, "q": 592\\
C: "input\_tokens": 592, "kv": 512, "q": 80
\end{convtool}
\begin{convagent}{Final choice: D}
``The simulator rounds usable prefix reuse down to 256-token units although the configured cache block size is 16; this can understate cache-affinity benefits and bias against C.''\par\smallskip
``D remains the predicted p95 winner (1.050s vs C 1.597s), despite its load-first routing not seeking cache affinity; recorded steps show this trade-off directly.''
\end{convagent}
\begin{conveval}
\convlabel{Evaluator only}\quad Reference answer: D. Outcome: upheld, kept.
\end{conveval}
\caption{GPT-5.6 Terra keeps load-only (D) on the upheld routing problem after weighing the lost credit against load. C is SGLang-inspired.}
\label{fig:case-routing-kept}
\end{figure}

\begin{figure}[t]
\convturn{First turn}
\begin{convuser}
\ldots\ Recommend one candidate for this workload.
\end{convuser}
\begin{convagent}{Initial choice: D}
``It has the lowest predicted measured-request p95 latency (1.0496 s).''
\end{convagent}
\convturn{Review turn}
\begin{convuser}
Review your recommendation before it is final. \ldots
\end{convuser}
\begin{convtool}{Tool calls 7, 10, and 11 of 23}
read \ldots/schedulers/chunked\_prefill.rs\\
search "PREFIX\_REUSE\_GRANULARITY\_TOKENS"\\
search "hit - hit \% PREFIX\_REUSE\_\ldots"
\end{convtool}
\begin{convtool}{Cited record of C, request 10}
"input\_tokens": 592, "kv": 512, "q": 80
\end{convtool}
\begin{convagent}{Final choice: C}
``The simulator undercredits cache hits at 512 tokens; with the intended 16-token blocks, C's cache-affine routing should retain substantially more of these repeated 592-token prompts, while D explicitly ignores reuse.''
\end{convagent}
\begin{conveval}
\convlabel{Evaluator only}\quad Reference answer: D. Outcome: upheld, abandoned.
\end{conveval}
\caption{In another repetition, GPT-5.6 Terra identifies the same rounding but switches to SGLang-inspired (C) without sizing the lost credit against the load gap.}
\label{fig:case-routing-left}
\end{figure}

\begin{figure}[!t]
\convturn{First turn}
\begin{convuser}
\ldots\ Recommend one candidate for this workload.
\end{convuser}
\begin{convtool}{Tool call 39 of 61}
read \ldots/schedulers/chunked\_prefill.rs:260-320
\convadd{+const PREFIX\_REUSE\_GRANULARITY\_TOKENS: u32 = 256;}
\convadd{+ hit - hit \% PREFIX\_REUSE\_GRANULARITY\_TOKENS}
\end{convtool}
\begin{convnote}{Narration}
``The 256-token reuse quantization may under-credit cache hits. I am checking whether that, or real load imbalance, drives the ranking.''
\end{convnote}
\begin{convagent}{Initial choice: A}
``\ldots\ The 256-token hit quantization under-credits versus block\_size 16, but observed hits are already 256-aligned and the gap is too large for that to reverse the ranking.''
\end{convagent}
\convturn{Review turn}
\begin{convuser}
Review your recommendation before it is final. \ldots
\end{convuser}
\begin{convtool}{Cited cost profile rows}
2,32,1,0,0.010049951553344727\\
2,64,1,0,0.05404348754882812
\end{convtool}
\begin{convagent}{Final choice: B}
``\ldots\ The scheduler instead quantizes hits to 256 and records kv=768/q=192, which sits on the profile cliff at 64 query tokens (\textasciitilde0.052s, same as a 960-token miss).''
\end{convagent}
\begin{conveval}
\convlabel{Evaluator only}\quad Reference answer: B. Outcome: overturned, corrected in review.
\end{conveval}
\caption{Grok 4.7 follows load-only (A) on the overturned routing problem, then sizes the lost credit with the cost profile and switches to SGLang-inspired (B).}
\label{fig:case-routing-corrected}
\end{figure}

\niparagraph{Conversations.}
\Cref{fig:case-routing-kept,fig:case-routing-left} show GPT-5.6 Terra identifying the same rounding in two repetitions of the upheld problem and deciding in opposite directions.
In the first, it compares one request across the records of load-only and SGLang-inspired and reads the lost credit as one side of a trade-off with load.
In the second, it cites only the rounded hit of SGLang-inspired and switches to it, although the reference simulator keeps load-only best by 75\%.
\Cref{fig:case-routing-corrected} shows Grok 4.7 finding the rounding in its first turn yet judging it too small to change the ranking.
In the review, it reads the cost profile, finds that the 192 uncredited tokens fall on a costly step size, and switches to the reference answer.

\subsection{Prefill-to-Decode Handoff with Independent Links}
\label{sec:case-handoff}

\begin{table*}[t]
\centering
\small
\begin{tabular}{@{}p{2.6cm}p{8.6cm}cc@{}}
\toprule
Design & Behavior & Upheld (s) & Overturned (s) \\
\midrule
Stay on source & Decodes on the prefill replica and never moves the KV cache. & 3.465 / 3.465 & 1.816 / 1.816 \\
Next instance & Moves the decode to the next replica above the prefill replica. & 3.771 / 3.771 & 2.424 / 2.424 \\
Least-active remote & Moves the decode to the remote replica with the fewest active calls, not counting transfers still in flight. & 3.769 / 3.769 & 2.425 / 2.425 \\
Least-outstanding & Counts active calls plus handoffs sent but not yet arrived, and moves the decode to the remote replica with the fewest. & 2.826 / 3.030 & 1.649 / 2.056 \\
\bottomrule
\end{tabular}
\caption{Handoff designs and their p95 end-to-end latency under the perturbed and reference simulators. The three remote placements follow disaggregated serving~\cite{distserve:osdi:2024}. The reference answer is least-outstanding on the upheld workload and stay on source on the overturned one.}
\label{tab:case-handoff}
\end{table*}

\niparagraph{Designs and perturbation.}
Each design decides where the follow-up requests of a shared document decode after prefill, and the target metric is p95 end-to-end latency.
\Cref{tab:case-handoff} describes the designs, and least-outstanding is the presented recommendation on both workloads.
In the modeled system, every KV transfer leaving a GPU shares its one 10\,GbE interface, whereas the perturbed simulator gives each ordered GPU pair its own link, so simultaneous transfers from one GPU finish together instead of queuing.
The perturbation replaces the switch topology with per-pair links, abridged below:
\begin{convtool}{Perturbation diff}
\convdel{- let switch = network\_sim.add\_device();}
\convdel{- for \&endpoint in physical\_endpoints.values() \{}
\convdel{-\ \ network\_sim.add\_link(endpoint, switch, ...)?;}
\convdel{-\ \ network\_sim.add\_link(switch, endpoint, ...)?;}
\convadd{+ for \&source in \&endpoints \{}
\convadd{+\ \ for \&destination in \&endpoints \{}
\convadd{+\ \ \ if source != destination \{}
\convadd{+\ \ \ \ network\_sim.add\_link(source, destination, ...)}
\end{convtool}

\niparagraph{Workloads and design-specific impact.}
Both workloads send five bursts in which three follow-ups per prefill replica move their KV caches together.
The overturned workload doubles the shared document and question to 2,560 and 384 tokens and halves the output to 3 tokens, so each moved transfer is about twice as large.
Only least-outstanding spreads decodes over seven replicas, so its simultaneous transfers leave one GPU for different destinations and never queue in the perturbed simulator.
Its mean first-token latency drops from 1.04 to 0.85\,s on the upheld workload and from 1.23 to 0.85\,s on the overturned one.
Stay on source sends no transfer, and the other two placements send each burst to the same few destinations, so their predictions do not change.
With the larger transfers, the corrected latency of least-outstanding rises past stay on source, which becomes best by 13\%, while on the upheld workload least-outstanding remains best by 14\%.

\niparagraph{Outcomes across models.}
Of the 39 perturbed conversations per workload, 37 keep the valid upheld recommendation and 29 replace the invalid overturned one, and 27 repetitions solve both problems.
Without simulator evidence, 16 and 26 are correct.

\begin{figure}[t]
\convturn{First turn}
\begin{convuser}
\ldots\ Recommend one candidate for this workload.
\end{convuser}
\begin{convtool}{Tool call 31 of 50}
read \ldots/systems/multi\_instance.rs:150-375
\convadd{+ network\_sim.add\_link(}
\convadd{+ \ \ source,}
\convadd{+ \ \ destination,}
\end{convtool}
\begin{convtool}{Cited system specification}
{\normalfont\scriptsize ``Each GPU has one 10 GbE network interface (1.25 GB/s) to the host network switch. All transfers leaving a GPU share its interface, and all transfers entering a GPU share its interface.''}
\end{convtool}
\begin{convtool}{Cited record of B}
"destination\_instance": 0, "kv\_tokens": 1632,\\ "request\_id": 29
\end{convtool}
\begin{convagent}{Initial choice: B}
``B counts in-flight handoffs and spreads decode load. Correcting the simulator's missing shared-source NIC contention raises its estimated p95 from 2.83s to about 3.03s \ldots''
\end{convagent}
\convturn{Review turn}
\begin{convuser}
Review your recommendation before it is final. \ldots
\end{convuser}
\begin{convagent}{Final choice: B}
``The simulator gives each GPU pair an independent link instead of sharing each GPU's 10GbE NIC across transfers. This understates B's transfer delay, but the correction does not reverse the ranking.''
\end{convagent}
\begin{conveval}
\convlabel{Evaluator only}\quad Reference answer: B, p95 3.030\,s. Outcome: upheld, kept.
\end{conveval}
\caption{GPT-6.1 Sol re-estimates least-outstanding (B) with the shared interface on the upheld handoff problem and keeps it ahead of stay on source (D).}
\label{fig:case-handoff-kept}
\end{figure}

\begin{figure}[t]
\convturn{First turn}
\begin{convuser}
\ldots\ Recommend one candidate for this workload.
\end{convuser}
\begin{convtool}{Tool call 21 of 72}
read \ldots/systems/multi\_instance.rs:175-260
\convadd{+ network\_sim.add\_link(}
\convadd{+ \ \ source,}
\convadd{+ \ \ destination,}
\end{convtool}
\begin{convagent}{Initial choice: B}
``Candidate B achieves the lowest 95th-percentile end-to-end latency (2.826s compared to D: 3.465s, A: 3.769s, and C: 3.771s).''
\end{convagent}
\convturn{Review turn}
\begin{convuser}
Review your recommendation before it is final. \ldots
\end{convuser}
\begin{convagent}{Final choice: D, with no review tool call}
``\ldots\ In the intended serving system, each GPU shares a single 10 GbE interface (1.25 GB/s), causing severe contention and serialization delays when multiple KV cache handoffs occur concurrently during arrival bursts. Candidate D avoids host network bottlenecks entirely by colocating decode execution on the prefill instance \ldots''
\end{convagent}
\begin{conveval}
\convlabel{Evaluator only}\quad Reference answer: B. Outcome: upheld, abandoned.
\end{conveval}
\caption{Gemini 3.8 Flash reaches the same lines on the upheld handoff problem but switches to stay on source (D) in the review on an unsized contention claim.}
\label{fig:case-handoff-left}
\end{figure}

\begin{figure}[t]
\convturn{First turn}
\begin{convuser}
\ldots\ Recommend one candidate for this workload.
\end{convuser}
\begin{convtool}{Tool call 29 of 64}
read \ldots/systems/multi\_instance.rs:179-250
\convadd{+ network\_sim.add\_link(}
\convadd{+ \ \ source,}
\convadd{+ \ \ destination,}
\end{convtool}
\begin{convagent}{Initial choice: D}
``\ldots\ By effectively load-balancing decodes across multiple replicas, Candidate D achieves the lowest 95th-percentile end-to-end latency of \textasciitilde1.649s.''
\end{convagent}
\convturn{Review turn}
\begin{convuser}
Review your recommendation before it is final. \ldots
\end{convuser}
\begin{convtool}{Tool call 8 of 68 and a cited record of D}
read \ldots/systems/multi\_instance.rs:190-230\\
"destination\_instance": 0, "kv\_tokens": 2912,\\ "request\_id": 11
\end{convtool}
\begin{convagent}{Final choice: C}
``Candidate D dispatches 3 requests simultaneously from each prefill replica. Due to the unmodeled per-GPU 10 GbE egress NIC bottleneck, those concurrent transfers would share 1.25 GB/s rather than each getting full bandwidth, tripling transfer latency from \textasciitilde0.186s to \textasciitilde0.558s. This adds \textasciitilde0.372s to Candidate D's p95 latency, pushing it to \textasciitilde2.02s and making Candidate C (1.816s, zero network transfers) the faster policy in reality.''
\end{convagent}
\begin{conveval}
\convlabel{Evaluator only}\quad Reference answer: C, with D at 2.056\,s. Outcome: overturned, corrected in review.
\end{conveval}
\caption{On the overturned problem, Gemini 3.8 Flash sizes the shared-interface delay of least-outstanding (D) in review and switches to stay on source (C).}
\label{fig:case-handoff-corrected}
\end{figure}

\niparagraph{Conversations.}
\Cref{fig:case-handoff-kept} shows GPT-6.1 Sol reaching the changed links in its first turn, checking them against the system specification, and re-estimating least-outstanding at about 3.03\,s, which matches the reference value of 3.030\,s and stays below stay on source.
\Cref{fig:case-handoff-left,fig:case-handoff-corrected} show Gemini 3.8 Flash identifying the same limitation on both workloads.
On the overturned workload, it sizes the added transfer delay at about 2.02\,s against the reference value of 2.056\,s and correctly switches to stay on source.
On the upheld workload, it makes no tool call in the review and switches to stay on source on a claim of severe contention, although the reference simulator keeps least-outstanding best by 14\%.
In both pairs, the same identification thus leads to opposite decisions, depending on whether the model sizes the limitation's effect.

\end{document}